\documentclass{article}
\usepackage{graphicx}
\usepackage{booktabs}
\usepackage[super]{cite}
\usepackage{amsmath,mathbbold,cases,amsfonts,amssymb,mathrsfs}
\usepackage{epsfig, multirow, subfigure, cite, txfonts}
\usepackage[small]{caption2}
\usepackage{color,hyperref}
\usepackage{fancyhdr}

\newtheorem{thm}{Theorem}[section]

\newtheorem{rem}[thm]{Remark}

\numberwithin{equation}{section}

\begin{document}
\title{A Robust Lane Detection Associated with Quaternion Hardy Filter}


\author{Wenshan Bi, Dong Cheng and Kit Ian Kou 
\thanks{This work was supported in part by the Science and Technology Development Fund, Macau SAR FDCT/085/2018/A2 and  the Guangdong Basic and Applied Basic Research Foundation (No. 2019A1515111185).}
\thanks{ W. Bi and K.I. Kou are with the Department of Mathematics, Faculty of Science and Technology, University of Macau, Macau, China (e-mail: wenshan0608@163.com; kikou@um.edu.mo).}
\thanks{D. Cheng is with the Research Center for Mathematics and Mathematics Education, Beijing Normal University at Zhuhai, 519087, China (e-mail: chengdong720@163.com).}
}

\maketitle


\begin{abstract}
In this article, a robust color-edge feature extraction method based on the Quaternion Hardy filter is proposed. The Quaternion Hardy filter is an emerging edge detection theory. It is along with the Poisson and conjugate Poisson smoothing kernels to handle various types of noise. Combining with the Quaternion Hardy filter, Jin's color gradient operator and Hough transform, the color-edge feature detection algorithm is proposed and applied to the lane marking detection. Experiments are presented to demonstrate the validity of the proposed algorithm. The results are accurate and robust with respect to the complex environment lane markings.
\end{abstract}
\text{Keywords:{Lane line detection, Hilbert transform, Quaternion Hardy filter, Color image. }}
\text{AMS Class Classification: 30E25, 11E88, 42B10}

A great deal of research has recently been conducted on intelligent vehicle systems in color image processing \cite{son2015real,ralston2016dynamic,luo2015video}. It consists of road lane detection, pedestrian avoidance, and obstacle detection. Among them, road lane detection is regarded as the most indispensable. The edge detector is considered to be the basic module for road lane detection. {Sangwine et al \cite{ell2014quaternion,sangwine1998colour} introduced a} well established color image edge detection algorithm.

In order to derive a robust and discriminative performance on lane marking, challenging scenarios such as shadows caused by illumination or weather changes are necessary. In 2014, Wu et al \cite{wu2014lane} introduced the lane-mark detector which is combining with Kalman filter and multi-adaptive thresholds method. Later on, Li et al \cite{li2016nighttime} in 2016 proposed the lane-mark detector based on Canny filter and Hough transform to extract lane boundaries or markings in complex scenarios. In order to provide robust and discriminative performance on lane marking, edge detectors which can handle various type of noises are necessary. Quaternion Hardy function \cite{hu2018phase} and Quaternion analytic signal \cite{hu2018phase} consist of Poisson and conjugate Poisson kernels. In 2016, Hu et al analyzed the Quaternion Hardy function \cite{hu2018phase} and Quaternion analytic signal \cite{hu2018phase} for edge detection problem.  It was proved that these edge detectors has significant properties associated with Quaternion Fourier transform. They can handle various types of noises and may be favorable for lane markings in complex scenarios.

Quaternion analysis recently draws wide attention in color image processing \cite{hu2018phase,  kou2017envelope, zou2016quaternion, kou2016uncertainty, yang2018edge}. The recently developed concepts of quaternion Fourier transform \cite{hitzer2016quaternion, hitzer2013orthogonal, hitzer2017general, cheng2019plancherel, hu2017quaternion}, quaternion representation \cite{zou2016quaternion} and quaternion analytic signal \cite{hu2018phase, kou2017envelope, bi2020robust}, which are based on quaternion algebra, have been found to be an indispensable tool in the representation of  the multidimensional signal. In earlier 1992, Ell \cite{ell1992hypercomplex} studied the two-sided quaternion Fourier transform and apply it to the color image analysis. In \cite{pei2001efficient}, Pei et al. gave a detailed interpretation of the quaternion Fourier transform. In \cite{hitzer2016quaternion, hitzer2013orthogonal, hitzer2017general}, Hitzer proposed the generalized quaternion Fourier transform by follow-up the works of Ell, Sanguine, and Ernst. As the generalization of the quaternion Fourier transform, quaternion linear canonical transform was firstly studied in Kou er al. \cite{kou2016uncertainty}. In 2017, Kou er al. \cite{kou2017envelope} studied the properties of quaternion linear canonical transform and applied it to envelop detection. In \cite{cheng2019plancherel}, Dong et al. give the Plancherel and inversion theorems of quaternion Fourier transform in the square-integrable signals space.
In 2016, Zou er al. \cite{zou2016quaternion} proposed two quaternion representation-based classification algorithms and applied them to color face recognition. Quaternion representation processes the color image in a holistic way and the structural information between color channels is preserved. Obviously, the experiments show the superiority of the proposed algorithm for color reconstruction and face recognition. In 2017, the quaternion analytic signal are studied and applied to edge detection \cite{hu2018phase}.

The goal of this study is introducing a novel lane-mark extraction algorithm which can handle complex scenarios. Firstly, edge enhancement based on Quaternion Hardy filter \cite{bi2020robust} was used to enhance the pre-processing lane image's edges. Then, the edges were detected by Jin's color detector \cite{jin2012improved} which is based on Di Zenzo's multichannel image gradient operator and combined with the straight lines which were detected by Hough transform \cite{ballard1981generalizing}. Moreover, an inside lane line extraction algorithm was proposed on the basis of slope constraint. Finally, the aim of marking the inside lane was realized. An overview flowchart of the proposed lane-mark extraction algorithm is shown in Fig. \ref{flowchart}. By experimenting with Canny lane markers, the proposed algorithm can realize the detection. Moreover, the approach can overcome the influence of uneven light and be able to eliminate the interference from the side lane line and the falling leaf, etc. Lane line detection is conducive to the vehicle running on its road. The contributions of this paper are summarized as follows.
\begin{itemize}
	\item[1.] A novel lane detector associated with Quaternion Hardy filter for color-edge road lane extraction is proposed and analyzed. Firstly, input the color road image, then edge enhancement based on Quaternion Hardy filter was applied to obtain the "smooth" lane images. Second, the edges were recorded by Jin's color gradient detector. Finally, combine with the straight lines obtained by Hough transform.
	\item[2.] We have carried out experimental research on the identification of road markings in various complex environments.
	Experimental results show the effectiveness and flexibility of the proposed method.
\end{itemize}

The key notations and acronyms used in this article are summarized in Table \ref{tab0}. The rest of this paper is organized as follows. Section \ref{QHF}  introduces some preliminaries of the quaternions, quaternion Fourier transform, and quaternion Hardy space. The definitions of  quaternion analytic signal and quaternion Hardy filter are recalled in Section \ref{QHF1}.
Section \ref{lane} introduces the proposed lane line detector, and the experimental results are presented in Section \ref{exp}.
Finally, concluding remarks are drawn in Section \ref{Con}.

\begin{table}[th]
	\caption{Key notations and acronyms used in this paper.}
    \centering
	\label{tab0}
	\begin{tabular}{c|ccccccc}
		\hline
		Notation  &Description  \\
		\hline	
		$\mathbb{R}^{2}$	    &2-dimensional Euclidean space  \\
		$\mathbb{H}$ 	        &quaternion space    \\
		$\mathbf{Q}^2(\mathbb{C}_{\mathbf{i} \mathbf{j}}^+)$	            &quaternion  Hardy space    \\							
		$\mathcal{F}_D[\cdot]$     &discrete quaternion Fourier transform  \\
		${\mathcal{F}}_D^{-1}[\cdot]$ &discrete inverse  quaternion Fourier transform\\
		$\mathcal{F}[\cdot]$    &quaternion Fourier transform\\
		${\mathcal{F}}^{-1}[\cdot]$ &inverse quaternion Fourier transform\\
		$f_q$   &quaternion analytic signal\\
		QHF	              &Quaternion Hardy filter	   \\
		\hline
	\end{tabular}
\end{table}
\section{Preliminaries}
\label{QHF}
In order to mention the proposed method, some basic knowledge about quaternion algebra \cite{kou2016uncertainty}, quaternion Fourier transform \cite{cheng2019plancherel, ell1992hypercomplex}, quaternion Hardy space \cite{hu2018phase}, Jin's gradient operator  \cite{jin2012improved} and Hough transform \cite{xu2014accurate} are reviewed as follows.
\subsection{Quaternion algebra and quaternion Fourier transform}
Let $\mathbb{H}$ denote the Hamiltonian skew field of quaternion
\begin{eqnarray}\label{H}
\mathbb{H}:=\{q=q_0+q_1\mathbf{i}+q_2\mathbf{j}+q_3\mathbf{k}| \ q_0, q_1, q_2, q_3\in \mathbb{R}\},
\end{eqnarray}
which is an associative anti-commutative four-dimensional algebra. A quaternion-valued function $f:\mathbb{R}^{2}\rightarrow\mathbb{H}$ can be represented by
\begin{eqnarray}\label{Eq.H-valued function}
f(x_{1},x_{2})=f_0(x_{1},x_{2})+f_1(x_{1},x_{2})\mathbf{i}+f_2(x_{1},x_{2})\mathbf{j}+f_3(x_{1},x_{2})\mathbf{k},
\end{eqnarray}
where $f_{n}:\mathbb{R}^{2}\rightarrow\mathbb{R}(n=0,1,2,3)$.
\subsection{Quaternion Hardy space}
Let $f\in L^1(\mathbb{R}^2, \mathbb{H})$ . {Then the quaternion Fourier transform (QFT) \cite{cheng2019plancherel, hitzer2007quaternion, hitzer2010directional} of $f$ is defined by}
\begin{equation}\label{Eq.2-sided QFT}
\mathcal{F}[f](w_1,w_2):=\frac{1}{2\pi}\int_{\mathbb{R}^{2}}e^{-\mathbf{i}w_1x_1}f(x_1,x_2)e^{-\mathbf{j}w_2x_2}d{x_1}d{x_2},
\end{equation}
where $w_l(l=1,2)$ denote the 2D angular frequencies. {Furthermore, if $f$ is an integrable $\mathbb{H}$-valued function defined on $\mathbb{R}^2$,  the  inverse quaternion Fourier transform (IQFT) \cite{cheng2019plancherel, hitzer2007quaternion, hitzer2010directional} of $f$ is defined by}
\begin{equation}\label{Eq.inverse 2-sided QFT}
{\mathcal{F}}^{-1}[f](x_1,x_2):=\frac{1}{2\pi}\int_{\mathbb{R}^{2}}e^{\mathbf{i}w_1x_1}f(w_1,w_2)e^{\mathbf{j}w_2x_2}d{w_1}d{w_2}.
\end{equation}

Let ${\mathbb{C}}^{+}:= \{z|z=x+si,
x, x \in \mathbb{R},s>0\}$, namely upper half complex plane and ${\mathbf{H}}^2(\mathbb{C}^{+})$ be the Hardy space on the upper half complex plane.
Let $ \mathbb{C}_{\mathbf{i} \mathbf{j}}:= \{( z_1, z_2)|z_1=x_1+s_1\mathbf{i}, z_2=x_2+s_2\mathbf{j},
x_l, s_l \in \mathbb{R}, l=1,2\}$. And a subset of $\mathbb{C}_{\mathbf{i} \mathbf{j}}$  is defined by $ \mathbb{C}_{\mathbf{i} \mathbf{j}}^{+}:= \{( z_1, z_2)|z_1=x_1+ s_1\mathbf{i}, z_2=x_2+ s_2\mathbf{j},x_l, s_l \in \mathbb{R}, s_l> 0, l=1,2\}.$
The quaternion Hardy space $\mathbf{Q}^2(\mathbb{C}_{\mathbf{i} \mathbf{j}}^+)$  consists of all functions $h$ satisfying \cite{Hu1},
\begin{eqnarray}\label{QHS conditions}
\left\{
\begin{aligned}
&\frac{\partial}{\partial \overline{z_{1}}}h(z_1, z_2)=0;\\
&h(z_1, z_2)\frac{\partial}{\partial \overline {z_{2}}}= 0;\\
& \left( \sup\limits_{\substack{s_1>0 \\
		s_2>0}}\int_{\mathbb{R}^2} |h(x_1+ s_1\mathbf{i}, x_2+s_2\mathbf{j} )|^{2}dx_{1}dx_{2} \right)^{\frac{1}{2}}  < \infty,
\end{aligned}\right.
\end{eqnarray}
where $\frac{\partial}{\partial \overline{z_{1}}}:=\frac{\partial}{\partial {x_{1}}}+\mathbf{i}\frac{\partial}{\partial {s_{1}}}$, $\frac{\partial}{\partial \overline{z_{2}}}:=\frac{\partial}{\partial {x_{2}}}+\mathbf{j}\frac{\partial}{\partial {s_{2}}}$.
\subsection{Jin's gradient operator}
Let $g$  be an $M\times{N}$ color image that maps a point $(x,y)$  to a vector $(g_1(x,y)$, $g_2(x,y)$, $g_3(x,y))$.
Specifically, for color image $g$, let  $(g_1(x,y)$, $g_2(x,y)$, $g_3(x,y))$ {denotes the red, the green, and the blue channel}, respectively.
Then the square of the variation of $f$ at the position $(x,y)$ with the distance $b$ in the direction $\theta$ is given by
\begin{eqnarray}\label{df2}
\begin{aligned}
dg^2& :=\|g(x+b{\cos\theta,y+b{\sin\theta})-g(x,y)}\|_2^2 \\
&\approx\sum\limits_{i=1}^{3}\left(\frac{\partial{g_i}}{\partial{x}}b\cos\theta+
\frac{\partial{g_i}}{\partial{y}}b\sin\theta\right)^2\\
&=b^2f(\theta),
\end{aligned}
\end{eqnarray}
where
\begin{eqnarray}\label{ftheta}
\begin{aligned}
g(\theta) :=&2\sum\limits_{i=1}^{3}\frac{\partial{g_i}}{\partial{x}}\frac{\partial{g_i}}{\partial{y}}\cos\theta\sin\theta\\
&+\sum\limits_{i=1}^{3}\left(\frac{\partial{g_i}}{\partial{x}}\right)^2\cos^2\theta+\sum\limits_{i=1}^{3}\left(\frac{\partial{g_i}}{\partial{y}}\right)^2\sin^2\theta.
\end{aligned}
\end{eqnarray}
Define
\begin{eqnarray}\label{ABC}
\left\{
\begin{aligned}
&H:= \sum\limits_{i=1}^{3}\left(\frac{\partial{g_i}}{\partial{x}}\right)^2; \\
&J:= \sum\limits_{i=1}^{3}\left(\frac{\partial{g_i}}{\partial{y}}\right)^2; \\
&K:= \sum\limits_{i=1}^{3}\frac{\partial{g_i}}{\partial{x}}\frac{\partial{g_i}}{\partial{y}}.
\end{aligned}\right.
\end{eqnarray}
Then the gradient magnitude $f_{\max}$ of the Jin's gradient operator is given by
\begin{eqnarray}\label{fMAX}
\begin{aligned}
g_{\max}(\theta_{\max}):&=\max_{0\leq \theta \leq 2 \pi}{g(\theta)}\\
=&\frac{1}{2}\bigg(H+K+\sqrt{(H-K)^2+(2J)^2}\bigg).
\end{aligned}
\end{eqnarray}
The gradient direction is defined as the value $\theta_{\max}$ that maximizes $g(\theta)$ over $0\leq \theta \leq 2 \pi$
\begin{eqnarray}\label{thetaMAX}
\theta_{\max}:=&\mbox{sgn}(J)\arcsin\bigg(\frac{g_{\max}-H}{2g_{\max}-H-K}\bigg),
\end{eqnarray}
where
$(H-K)^2+J^2\neq0$,
$\mbox{sgn}(J)=\left\{
\begin{array}{ll}
1, & {J\geq0;} \\
-1, & {J<0.}
\end{array}
\right.$
While $(H-K)^2+J^2=0$, $\nonumber\theta_{\max}$ is undefined.
\subsection{Hough transform}

{The Hough transform} is done on a binary image, obtained after processing the original image by an edge detector \cite{illingworth1988survey, xu2014accurate, popplewell2014multispectral}. This is  an ingenious method that converts such global curve detection problem into an efficient peak detection problem in parameter space.
	In addition, {Dorst et al} \cite{dorstguided, dorst1986best, dorst1984discrete, dorst1984spirograph} studied the application of plane-based geometric algebra in the representation of straight lines.
The basic idea of Hough transform is to use the duality relationship between the points and lines in the image domain and the parameter domain. In particular, the slope-intercept form is a classical parametric form for straight line detection, as is shown in Eq. (\ref{Eq.HF0}):
\begin{eqnarray}\label{Eq.HF0}
f(x,y)=y-kx-b=0,
\end{eqnarray}
where $(x,y)$ is the coordinate of the pixel being mapped, and the parameters $k$ and $b$ are the slop and y-intercept of the line respectively.  It's worth noting that this method is sensitive to the choice of coordinate axes on the image plane because both the parameters become unbounded when $k$ is infinite.

Duda and Hart \cite{duda1972use} lucidly resolved the issue of unboundedness by mapping a point in image space to a sinusoidal curve in $\rho-\theta$ parameter space. Here we use HoughlinesP method to detect lines, which is based on statistics not only could detect the two endpoints of the straight lines but also have the advantage of high efficiency. $\rho$ and $\theta$ satisfy the following relationship:
\begin{eqnarray}\label{Eq.HF1}
\rho(\theta)=x\cos{\theta}+y\sin{\theta},
\end{eqnarray}
where $\theta$ is the angle between the line and the $Y$-axis, $\rho$ is the distance from the origin to the line in the image coordinate, $\rho >0$ and $\rho(\theta+\pi)=-\rho(\theta)$.

The first step of Hough transform is initialize accumulator to all zeros, followed by the parameter space is quantized in intervals of $\Delta \rho$ and $\Delta \theta$ and corresponding accumulators are created to collect the evidence (or vote) of object pixels satisfying Eq.(\ref{Eq.HF1}).  Consider the ($m$, $n$)th accumulator corresponding to intervals [($m-1)\Delta \rho$, $m \Delta \rho$) and [($n-1)\Delta \theta$, $n \Delta \theta$ ) where $m,n\in \mathcal{Z}^+$ . For each object pixel satisfying Eq.(\ref{Eq.HF1}) with parameters in the above range, the vote in the ($m$, $n$)th accumulator is incremented by 1. In the peak detection phase, the accumulators having number of votes above a critical threshold correspond to straight lines in the image. Therefore, the line detection problem is converted to peak  statistics problems.
\section{Quaternion Hardy filter}\label{QHF1}
The analytic signal \cite{garnett2007bounded} of a given real signal $f$ is defined by
\begin{equation}\label{Eq.1D analytic signal}
f_a(x) :=f(x)+i\mathcal{H}[f](x),   \; \; x\in\mathbb{R},
\end{equation}
where $\mathcal{H}[f]$ denotes the Hilbert transform  \cite{king2009hilbert} of $f$. By a direct computation, the Fourier transform of  $f_a$ is given by
\begin{equation}\label{fa}
\widehat{f_a}(w)=\left(1+\mbox{sgn}(w)\right)\widehat{f}(w), \; \; w\in\mathbb{R}.
\end{equation}
\begin{figure}[t]\label{scale}
	\centering
	\includegraphics[height=7.2cm,width=0.75\textwidth]{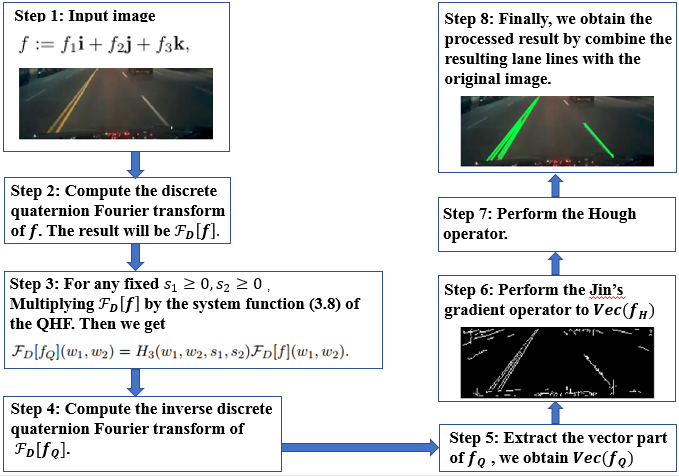}
	\caption{The flowchart of the proposed lane detection method.}
	\label{flowchart}
\end{figure}
The analytic signal of $f$ can be regarded as the output signal of a filter with input $f$. The system function of this filter is given by
\begin{equation}\label{filter1}
H_1(w):=1+\mbox{sgn}(w).
\end{equation}
\noindent Motivated by the success of analytic signal in the real signal processing, we try to use the alternative tool in the 2D  signal processing.

Give a 2D quaternion-valued signal $f$. The quaternion analytic signal \cite{bulow2001hypercomplex} $f_q$ is defined by

\begin{equation}\label{QanalyticS}
f_q(x_1,x_2):=f(x_1,x_2)+\mathbf{i}\mathcal{H}_{x_1}[f](x_1)+\mathcal{H}_{x_2}[f](x_2)\mathbf{j}+\mathbf{i}\mathcal{H}_{{x_1}{x_2}}[f](x_1,x_2)\mathbf{j},
\end{equation}
where $\mathcal{H}_{x_1}$, $\mathcal{H}_{x_2}$ are the partial Hilbert transforms with respect to $x_1,x_2$ respectively, while  $\mathcal{H}_{{x_1}{x_2}}$  the 2D Hilbert transform. By a direct computation \cite{Kou2}, the quaternion Fourier transform of $f_q$ is written as
\begin{equation}\label{fq}
\mathcal{F}[{f_q}](w_1,w_2)=[1+\mbox{sgn}(w_1)][1+\mbox{sgn}(w_2)]\mathcal{F}[{f}](w_1,w_2).
\end{equation}
The quaternion analytic signal of $f$ can be regarded as the output signal of a filter with input $f$.
The system function of this filter is given by
\begin{equation}\label{filter2}
H_2(w_1,w_2):=[1+\mbox{sgn}(w_1)][1+\mbox{sgn}(w_2)].
\end{equation}
By  the inverse quaternion Fourier transform on the both side of (\ref{fq}), we obtain
\begin{equation}\label{fq0}
f_q(x_1,x_2)=\mathcal{F}^{-1}[H_2\mathcal{F}[{f}]](x_1,x_2).
\end{equation}
In this paper, we apply quaternion Hardy filter as a lane detector of color image. The system function of quaternion Hardy filter is defined by
\begin{equation}\label{filter3}
\begin{aligned}
H_3(w_1,w_2,s_1,s_2)
:=&e^{-\mid{w_1}\mid{s_1}}e^{-\mid{w_2}\mid{s_2}}H_2(w_1,w_2)\\
=&e^{-\mid{w_1}\mid{s_1}}e^{-\mid{w_2}\mid{s_2}}[1+\mbox{sgn}(w_1)][1+\mbox{sgn}(w_2)].
\end{aligned}
\end{equation}

Based on the above analysis, we have the following theorem.
\begin{figure}[t]
	\centering \includegraphics[height=7.5cm,width=0.75\textwidth]{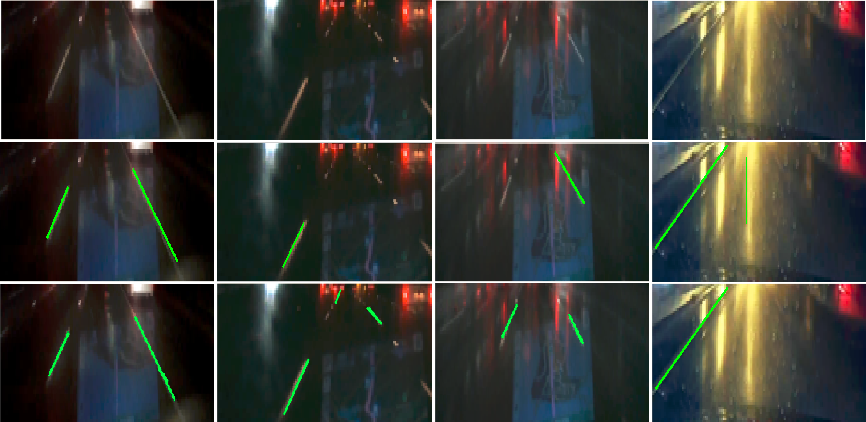}
	\caption{The first row is the original image in rainy night driving condition. The second row  and the third row are the lane detection results which are captured by Jin's algorithm and the the proposed algorithm, respectively.}
	\label{Nrain}
\end{figure}
\begin{thm}\label{theorem}
	Let $f\in L^2(\mathbb{R}^2, \mathbb{H})$, and
	\begin{eqnarray}\label{fQ}
	f_Q(x_1,x_2,s_1,s_2)={\mathcal{F}}^{-1}[H_3\mathcal{F}[f]](x_{1},x_{2},s_1,s_2).
	\end{eqnarray}
	Where $H_3$ is given by (\ref{filter3}).
	Then $f_Q\in\mathbf{Q}^2(\mathbb{C}_{\mathbf{i} \mathbf{j}}^+)$.
\end{thm}
The proof of Theorem \ref{theorem} can be found in \cite{QHF}. It states that the output signal of the quaternion Hardy filter belongs to the quaternion Hardy space $\mathbf{Q}^2(\mathbb{C}_{\mathbf{i} \mathbf{j}}^+)$.
\begin{rem}
	 {The quaternion analytic signal} is a special case of quaternion Hardy filter. That is, when $s_1=s_2=0$ in Theorem \ref{theorem}, the system function $H_3(x_1, x_2, 0, 0)$ reduces to $H_2(w_1, w_2)$, and the signal $f_Q(x_{1}, x_{2}, 0, 0)$ reduces to $f_q(x_1, x_2)$ which is defined by equation (\ref{fq0}).
\end{rem}
\begin{rem}
		In image processing, quaternion Hardy filter can not only suppress the high frequency of image, but also enhance the edge feature. It's worth noting that, the system function $H_3(x_1, x_2, s_1, s_2)$ plays a key role in quaternion Hardy filter.
		It has the advantage of both Hilbert transform and low-pass filter. On the one hand, the factor $[1+\mbox{sgn}(w_1)][1+\mbox{sgn}(w_2)]$ in $H_3(x_1, x_2, s_1, s_2)$  performs Hilbert transform on the input signal. It can enhance the edge feature. On the other hand, $e^{-\mid{w_1}\mid{s_1}}e^{-\mid{w_2}\mid{s_2}}$ could improve the ability of noise immunity for the quaternion Hardy filter.
\end{rem}
\section{The proposed lane detection algorithm}\label{lane}
The flowchart is shown in Fig. \ref{flowchart}.
In the following, let's  consider the details of the proposed algorithm. They are divided by the following steps.
\begin{itemize}
	\item []
	{\bf{Step} 1}. Input the road image $f(x_1,x_2)$. By quaternion representation, it can be written as
	\begin{eqnarray}\label{fD}
	f(x_1,x_2)=f_R(x_1,x_2)\mathbf{i}+f_G(x_1,x_2)\mathbf{j}+f_B(x_1,x_2)\mathbf{k},
	\end{eqnarray}
	where $f_R,f_G$ and $f_B$ represent the red, green and blue components of color image $f$, respectively.
	
	\item []
	{\bf{Step} 2}. Compute the discrete quaternion Fourier transform (DQFT) \cite{sangwine1997discrete}, $\mathcal{F}_D[f](w_1,w_2)$, of the image $f$ from Step 1, we have
	\begin{eqnarray}\label{fD}
	\mathcal{F}_D[f](w_1,w_2)=\sum_{x_1=0}^{M-1}\sum_{x_2=0}^{M-1}e^{-\mathbf{i}2{\pi}(w_1x_1/M)}f(x_1,x_2)e^{-\mathbf{j}2{\pi}(w_2x_2/M)}.
	\end{eqnarray}
	\begin{figure}[t]
		\centering
		\includegraphics[height=7.5cm,width=0.75\textwidth]{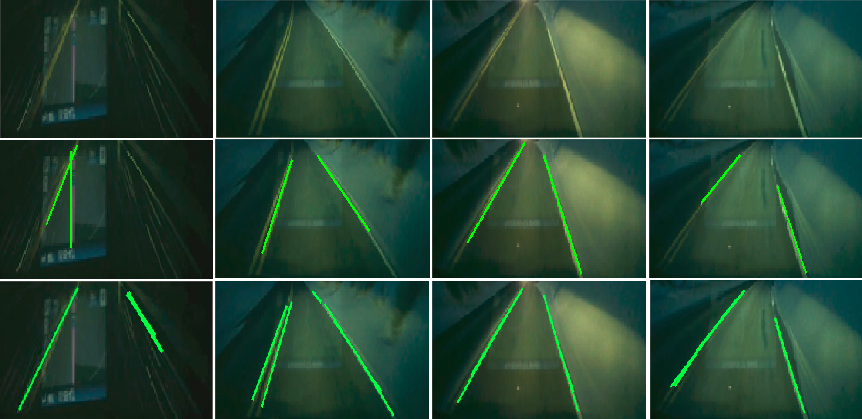}
			\caption{The first row is the original image in snowy night driving condition. The second row  and the third row are the lane detection results which are captured by Jin's algorithm and the the proposed algorithm, respectively.}
		\label{Nsnow}
	\end{figure}
	\item []
	{\bf{Step} 3}. For the fixed positive $s_1$ and $s_2$, calculate $H_3$ by Eq. (\ref{filter3}), we have
	\begin{equation*}
		H_3(w_1,w_2,s_1,s_2):=e^{-\mid{w_1}\mid{s_1}}e^{-\mid{w_2}\mid{s_2}} [1+\mbox{sgn}(w_1)][1+\mbox{sgn}(w_2)].
	\end{equation*}
	
	Multiplying $\mathcal{F}_D[f]$ with $H_3$, from Step 2 and 3, respectively.  By Theorem \ref{theorem}, we have
	\begin{eqnarray}\label{FfQ}
	\mathcal{F}_D[f_Q](w_1,w_2)= H_3(w_1,w_2,s_1,s_2) \mathcal{F}_D[f](w_1,w_2).
	\end{eqnarray}

	\item []
	{\bf{Step} 4}. Compute the inverse DQFT of  $\mathcal{F}_D[f_Q]$, from Step 3,
	\begin{eqnarray}\label{IDQFT}
	\begin{aligned}
	f_Q(x_1,x_2)
	=&\sum_{w_1=0}^{M-1}\sum_{w_2=0}^{M-1}e^{\mathbf{i}2{\pi}(w_1x_1/M)}\mathcal{F}_D[f_Q](w_1,w_2)e^{\mathbf{j}2{\pi}(w_2x_2/M)}\\
	=&\sum_{w_1=0}^{M-1}\sum_{w_2=0}^{M-1}e^{\mathbf{i}2{\pi}(w_1x_1/M)}[1+\mbox{sgn}(w_1)][1+\mbox{sgn}(w_2)]e^{-{3}\mid{w_1}\mid}e^{-{20}\mid{w_2}\mid}\\
	&\mathcal{F}_D[f](w_1,w_2)e^{\mathbf{j}2{\pi}(w_2x_2/M)}.
	\end{aligned}
	\end{eqnarray}
	We obtain $f_Q$.
	
	\item []
	{\bf{Step} 5}.  Extract the vector part of $f_Q$, we obtain
	\begin{eqnarray}
	\mathbf{Vec}(f_Q):=h_1\mathbf{i}+h_2\mathbf{j}+h_3\mathbf{k},
	\end{eqnarray}
	where $h_k$, $k=1,2,3$ are real-valued functions.
	
	\item []
	{\bf{Step} 6}. Applying the Jin's gradient operator \cite{jin2012improved} to $\mathbf{Vec}(f_Q)$.
	
	\item []
	{\bf{Step} 7}. Furthermore, perform the Hough operator \cite{ballard1981generalizing}.
	
	\item []
	{\bf{Step} 8}. Finally, we show the final result by placing the resulting lane lines on the original image.
	\label{algorithm}
\end{itemize}
\section{Experiment}
\label{exp}
To verify the effectiveness of quaternion hardy filter-based lane detection algorithm, we conduct experiments on 27 video clips.
{The experiments are programmed in Matlab R2016b.
The proposed algorithm is compared with previous work \cite{son2015real}, \cite{aly2008real}, \cite{jung2015efficient}, \cite{Yoo2013gradient}, \cite{liu2013lane}, \cite{lee2018robust}, and \cite{jin2012improved}. Here, Jin \cite{jin2012improved} means that the proposed method without steps 2-4 and provides a baseline reference results. } The previous work involved six different approaches that are typical of the field of lane line detection. \\
The video clips used in the experiment were from public dataset {\cite{dataset}}, which were shot using dashboard cameras in the US and Korea.  This data set contains a total of 49 video clips, and 27 are randomly selected for this experiment. The 27 video clips consist of 44159 frame images which divided into 6 different weather conditions: day-clear (5), day-rainy (5), day-snowy (5), night-clear (5), night-rainy (5) and night-snowy (2). The number in parentheses after the weather label represent the number of video clips used in the experiment.
\subsection{Visual results}
Visual results including rain night and snow night are illustrate in Fig. \ref{Nrain} and Fig. \ref{Nsnow}, respectively. When car lights form bright lines on the road on a rainy night, the comparison method may fail.
It affects the detection of lane lines.
{For example, in the second row and fourth column of Fig. 2, Jin's method is obviously affected by the light, thus making a misjudgment of the lane line.
In addition, Jin's method (e.g., the second row and first column in Fig. 3) detects the boundary line of the navigation screen and ignores the actual lane line on the right.}
Gradient-based approaches \cite{Yoo2013gradient} may preserve the boundary portion of the navigation screen while ignoring the actual lane lines when they are  reflective navigation screens are present.
In general, the proposed method can achieve satisfactory results under complex weather conditions and illumination conditions.
{The proposed method cannot, however, detect lane markings when affected by surrounding buildings or  windshield wipers, as shown in Fig. 4.}
\subsection{Quantitative comparisons}
For each image, two boundary lines are drawn on the long edge of the lane mark in the presence of the lane mark, and the real value of the ground is obtained manually. The detection rate is obtained by the following
\begin{eqnarray}\label{rate}
DR=\frac{correct}{total}\times100\%
\end{eqnarray}
where $correct$ represents the number of images correctly detected and $total$ represents the total number of images participated in the detection.

\begin{figure}[t]
	\centering
	\includegraphics[height=7.5cm,width=0.75\textwidth]{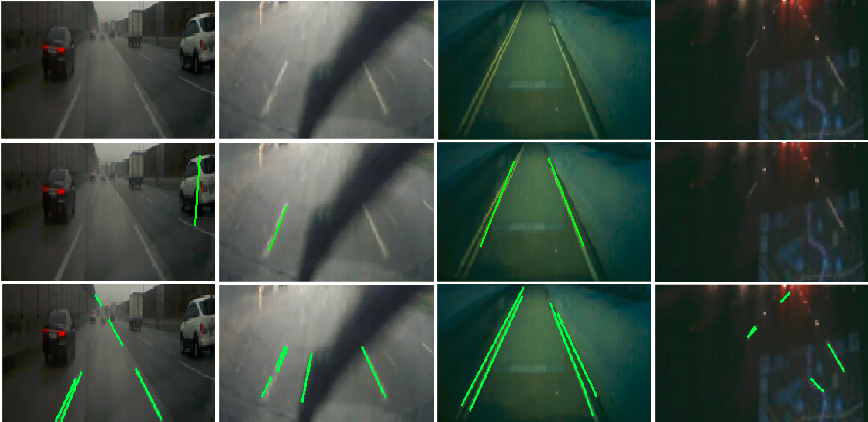}
		\caption{The first row is the original image in complex driving condition. The second row  and the third row are the false lane detection results which are captured by Jin's algorithm and the the proposed algorithm, respectively.}
	\label{wrong}
\end{figure}
We present a quantitative evaluation for seven methods in Table \ref{tab1}. It can be seen from Table \ref{tab1} that all  the algorithms can generate good results when lane images are captured with good conditions, as shown in clear days and nights.
For clear day sets, QHF-based method achieves a 99.4\% accuracy among the compared methods. For rain day and snow day sets, the proposed method keeps steady in the accuracy.
On clear night sets, the accuracy of the proposed method is still in the Top three. In the rainy night data set, the accuracy of the proposed method was still 88.5\%, despite the fact that raindrops falling on the windshield caused considerable interference with the lane lines in the image. For the snowy night data set, the other six methods were not considered, and the proposed method was accurate to 95.4\%.
We have to say that there are some situations that are hard to dealt with. Fig. \ref{wrong} gives several failure samples. For those images affected by wipers, illuminations and weather conditions, our method faces challenges. A potential solution to solve this problem is to use $\Lambda$ ROI (region of interest) \cite{lee2018robust}. However, ROI-based lane detection method require much computation cost. Therefore, more optimization works needs to be done in the future.
\begin{table}[th]
\caption{Comparison of detection rate of different algorithms for color image lane detection on 44159 images .}
	\label{tab1}
\resizebox{\textwidth}{!}{ %
\begin{tabular}{c|cccccccc}
		\hline
		DR (\%)  &Son\cite{son2015real}  &Aly\cite{aly2008real}   &Jung\cite{jung2015efficient}   &Yoo\cite{Yoo2013gradient}   &Liu\cite{liu2013lane}   &Lee\cite{lee2018robust} &Jin\cite{jin2012improved} &Ours\\
		\hline	
		Clear-Day	          	&93.1 &96.3   &88$\sim$98  &96.36 &-      &99     &99.3     &\textbf{99.4}\\
		Rain-Day	            &89   &-    &-   &96.19      &\textbf{98.67}      &96.9   &96.8     &97\\
		Snow-Day	            &-    &-    &-	    &-     &-      &93    &94.2      &\textbf{95.5}\\
		\hline								
		Clear-Night          	&94.3 &-  &87.7   &96.5  &94.06      &\textbf{98}	& 95.7       &96.1\\
		Rain-Night	          	&93   &-  &-   &-     &-      &\textbf{93.6}  &79.4     &88.5\\
		Snow-Night	            &-	  &-   &- &-      &-      &- &89.9	 &\textbf{95.4}\\
		\hline
		Processing\\
		time/ & 33 &20 &3.9 &50 &200 &35.4 &300  &800\\
		frame(ms) \\
		\hline
		Platform  &PC &PC &PC &PC &PC &ARM &PC &PC \\
		CPU &Unknow &Core 2 &2.8GHz &Unknow &2.3GHz &A9 GHz &2.3GHz &2.3GHz\\
		& & @2.4GHz & & & &@800MHz\\
\bottomrule
    \end{tabular}}
\end{table}
\section{Conclusion and future work}
\label{Con}
In this paper, we proposed a QHF-based lane detection method for color image. It exploits quaternion Hardy filter to generate clear feature image, and it use Jin's
gradient operator and Hough transform to provided lane lines, which ensures accuracy lane lines.
Our method could handle with multiple situations such as various weather conditions, illuminations and lane markings. Experiments conducted on  the dataset indicate that the proposed method achieves satisfactory results.
The results were accurate and robust with respect to the complex environment lane markings.
Finally, we are interested in extending our proposed quaternion lane line detection idea to color video for lane line detection in the future.
{In order to optimize the algorithm, we will also consider fast algorithms \cite{hitzer2013orthigonal} and algorithm optimization software \cite{GAALOP}.}

%




\section*{Acknowledgment}
This research was supported by Macao Science and Technology Development Fund (No. FDCT/085/2018/A2).

\bibliographystyle{unsrt}
\bibliography{arxiv0806lane}

\begin{thebibliography}{10}

\bibitem{son2015real}
Jongin Son, Hunjae Yoo, Sanghoon Kim, and Kwanghoon Sohn.
\newblock Real-time illumination invariant lane detection for lane departure
  warning system.
\newblock {\em Expert Systems with Applications}, 42(4):1816--1824, 2015.

\bibitem{ralston2016dynamic}
Robert~E Ralston and Justin~D Bloom.
\newblock Dynamic routing intelligent vehicle enhancement system, September~6
  2016.
\newblock US Patent 9,435,652.

\bibitem{luo2015video}
Yun Luo and Dieter Hoetzer.
\newblock Video based intelligent vehicle control system, October~20 2015.
\newblock US Patent 9,165,468.

\bibitem{ell2014quaternion}
Todd~A Ell, Nicolas Le~Bihan, and Stephen~J Sangwine.
\newblock {\em Quaternion Fourier transforms for signal and image processing}.
\newblock John Wiley \& Sons, 2014.

\bibitem{sangwine1998colour}
Stephen~John Sangwine.
\newblock Colour image edge detector based on quaternion convolution.
\newblock {\em Electronics Letters}, 34(10):969--971, 1998.

\bibitem{wu2014lane}
Pei-Chen Wu, Chin-Yu Chang, and Chang~Hong Lin.
\newblock Lane-mark extraction for automobiles under complex conditions.
\newblock {\em Pattern Recognition}, 47(8):2756--2767, 2014.

\bibitem{li2016nighttime}
Yadi Li, Liguo Chen, Haibo Huang, Xiangpeng Li, Wenkui Xu, Liang Zheng, and
  Jiaqi Huang.
\newblock Nighttime lane markings recognition based on canny detection and
  hough transform.
\newblock In {\em 2016 IEEE International Conference on Real-time Computing and
  Robotics (RCAR)}, pages 411--415. IEEE, 2016.

\bibitem{hu2018phase}
Xiao-Xiao Hu and Kit~Ian Kou.
\newblock Phase-based edge detection algorithms.
\newblock {\em Mathematical Methods in the Applied Sciences},
  41(11):4148--4169, 2018.

\bibitem{kou2017envelope}
Kit~Ian Kou, Ming-Sheng Liu, Jo{\~a}o~Pedro Morais, and Cuiming Zou.
\newblock Envelope detection using generalized analytic signal in 2d qlct
  domains.
\newblock {\em Multidimensional Systems and Signal Processing},
  28(4):1343--1366, 2017.

\bibitem{zou2016quaternion}
Cuiming Zou, Kit~Ian Kou, and Yulong Wang.
\newblock Quaternion collaborative and sparse representation with application
  to color face recognition.
\newblock {\em IEEE Transactions on image processing}, 25(7):3287--3302, 2016.

\bibitem{kou2016uncertainty}
Kit~Ian Kou, Jianyu Ou, and Joao Morais.
\newblock Uncertainty principles associated with quaternionic linear canonical
  transforms.
\newblock {\em Mathematical Methods in the Applied Sciences},
  39(10):2722--2736, 2016.

\bibitem{yang2018edge}
Yan Yang, Kit~Ian Kou, and Cuiming Zou.
\newblock Edge detection methods based on modified differential phase
  congruency of monogenic signal.
\newblock {\em Multidimensional Systems and Signal Processing}, 29(1):339--359,
  2018.

\bibitem{hitzer2016quaternion}
Eckhard Hitzer.
\newblock The quaternion domain fourier transform and its properties.
\newblock {\em Advances in Applied Clifford Algebras}, 26(3):969--984, 2016.

\bibitem{hitzer2013orthogonal}
Eckhard Hitzer and Stephen~J Sangwine.
\newblock The orthogonal 2d planes split of quaternions and steerable
  quaternion fourier transformations.
\newblock In {\em Quaternion and Clifford Fourier transforms and wavelets},
  pages 15--39. Springer, 2013.

\bibitem{hitzer2017general}
Eckhard Hitzer.
\newblock General two-sided quaternion fourier transform, convolution and
  mustard convolution.
\newblock {\em Advances in Applied Clifford Algebras}, 27(1):381--395, 2017.

\bibitem{cheng2019plancherel}
Cheng Dong and Kou~Kit Ian.
\newblock Plancherel theorem and quaternion fourier transform for square
  integrable functions.
\newblock {\em complex variables and elliptic equations}, 64(2):223--242, 2019.

\bibitem{hu2017quaternion}
Xiao-Xiao Hu and Kit~Ian Kou.
\newblock Quaternion fourier and linear canonical inversion theorems.
\newblock {\em Mathematical Methods in the Applied Sciences}, 40(7):2421--2440,
  2017.

\bibitem{bi2020robust}
Wenshan Bi, Dong Cheng, and Kit~Ian Kou.
\newblock A robust color edge detection algorithm based on quaternion hardy
  filter.
\newblock {\em arXiv preprint arXiv:2001.01800}, 2020.

\bibitem{ell1992hypercomplex}
Todd~Anthony Ell.
\newblock {\em Hypercomplex spectral transformations}.
\newblock PhD thesis, University of Minnesota, 1992.

\bibitem{pei2001efficient}
Soo-Chang Pei, Jian-Jiun Ding, and Ja-Han Chang.
\newblock Efficient implementation of quaternion fourier transform,
  convolution, and correlation by 2-d complex fft.
\newblock {\em IEEE Transactions on Signal Processing}, 49(11):2783--2797,
  2001.

\bibitem{jin2012improved}
Lianghai Jin, Hong Liu, Xiangyang Xu, and Enmin Song.
\newblock Improved direction estimation for di zenzo's multichannel image
  gradient operator.
\newblock {\em Pattern recognition}, 45(12):4300--4311, 2012.

\bibitem{ballard1981generalizing}
Dana~H Ballard.
\newblock Generalizing the hough transform to detect arbitrary shapes.
\newblock {\em Pattern recognition}, 13(2):111--122, 1981.

\bibitem{xu2014accurate}
Zezhong Xu, Bok-Suk Shin, and Reinhard Klette.
\newblock Accurate and robust line segment extraction using minimum entropy
  with hough transform.
\newblock {\em IEEE Transactions on Image Processing}, 24(3):813--822, 2014.

\bibitem{hitzer2007quaternion}
Eckhard~MS Hitzer.
\newblock Quaternion fourier transform on quaternion fields and
  generalizations.
\newblock {\em Advances in Applied Clifford Algebras}, 17(3):497--517, 2007.

\bibitem{hitzer2010directional}
Eckhard~MS Hitzer.
\newblock Directional uncertainty principle for quaternion fourier transform.
\newblock {\em Advances in Applied Clifford Algebras}, 20(2):271--284, 2010.

\bibitem{illingworth1988survey}
John Illingworth and Josef Kittler.
\newblock A survey of the hough transform.
\newblock {\em Computer vision, graphics, and image processing}, 44(1):87--116,
  1988.

\bibitem{popplewell2014multispectral}
Khary Popplewell, Kaushik Roy, Foysal Ahmad, and Joseph Shelton.
\newblock Multispectral iris recognition utilizing hough transform and modified
  lbp.
\newblock In {\em 2014 IEEE International Conference on Systems, Man, and
  Cybernetics (SMC)}, pages 1396--1399. IEEE, 2014.

\bibitem{dorstguided}
Leo Dorst.
\newblock A guided tour to the plane-based geometric algebra pga.

\bibitem{dorst1986best}
Leo Dorst and Arnold~WM Smeulders.
\newblock Best linear unbiased estimators for properties of digitized straight
  lines.
\newblock {\em IEEE transactions on pattern analysis and machine intelligence},
  (2):276--282, 1986.

\bibitem{dorst1984discrete}
Leo Dorst and Arnold~WM Smeulders.
\newblock Discrete representation of straight lines.
\newblock {\em IEEE Transactions on Pattern Analysis and Machine Intelligence},
  (4):450--463, 1984.

\bibitem{dorst1984spirograph}
Leo Dorst and Robert~PW Duin.
\newblock Spirograph theory: A framework for calculations on digitized straight
  lines.
\newblock {\em IEEE transactions on pattern analysis and machine intelligence},
  (5):632--639, 1984.

\bibitem{duda1972use}
Richard~O Duda and Peter~E Hart.
\newblock Use of the hough transformation to detect lines and curves in
  pictures.
\newblock {\em Communications of the ACM}, 15(1):11--15, 1972.

\bibitem{garnett2007bounded}
John Garnett.
\newblock {\em Bounded analytic functions}, volume 236.
\newblock Springer Science \& Business Media, 2007.

\bibitem{king2009hilbert}
Frederick~W King.
\newblock {\em Hilbert transforms}, volume~1.
\newblock Cambridge University Press Cambridge, 2009.

\bibitem{bulow2001hypercomplex}
Thomas Bulow and Gerald Sommer.
\newblock Hypercomplex signals-a novel extension of the analytic signal to the
  multidimensional case.
\newblock {\em IEEE Transactions on signal processing}, 49(11):2844--2852,
  2001.

\bibitem{sangwine1997discrete}
Stephen~John Sangwine.
\newblock The discrete quaternion fourier transform.
\newblock 1997.

\bibitem{aly2008real}
Mohamed Aly.
\newblock Real time detection of lane markers in urban streets.
\newblock In {\em 2008 IEEE Intelligent Vehicles Symposium}, pages 7--12. IEEE,
  2008.

\bibitem{jung2015efficient}
Soonhong Jung, Junsic Youn, and Sanghoon Sull.
\newblock Efficient lane detection based on spatiotemporal images.
\newblock {\em IEEE Transactions on Intelligent Transportation Systems},
  17(1):289--295, 2015.

\bibitem{Yoo2013gradient}
Hunjae Yoo, Ukil Yang, and Kwanghoon Sohn.
\newblock Gradient-enhancing conversion for illumination-robust lane detection.
\newblock {\em IEEE Transactions on Intelligent Transportation Systems},
  14(3):1083--1094, 2013.

\bibitem{liu2013lane}
Guorong Liu, Shutao Li, and Weirong Liu.
\newblock Lane detection algorithm based on local feature extraction.
\newblock In {\em 2013 Chinese Automation Congress}, pages 59--64. IEEE, 2013.

\bibitem{lee2018robust}
Lee Chanho and Ji-Hyun Moon.
\newblock Robust lane detection and tracking for real-time applications.
\newblock {\em IEEE Transactions on Intelligent Transportation Systems},
  19(12):4043--4048, 2018.

\bibitem{dataset}
\url{https://drive.google.com/file/d/1315Ry7isciL3nRvU5SCXM_-4meR2MyI/view?usp=sharing}.

\bibitem{hitzer2013orthigonal}
Eckhard Hitzer and J.~Sangwine Stephen.
\newblock The orthogonal 2d planes split of quaternions and steerable
  quaternion fourier transformations.
\newblock {\em Quaternion and Clifford Fourier transforms and wavelets}, pages
  15--39, 2013.

\bibitem{GAALOP}
Gaalop.
\newblock \url{http://www.gaalop.de/dhilden/}.

\end{thebibliography}

\end{document}